\documentclass[letterpaper]{article}
\usepackage{aaai}
\usepackage{times}
\usepackage{helvet}
\usepackage{courier}

\usepackage{caption}
\usepackage{subcaption}
\usepackage{bbold}
\usepackage{amsmath}

\usepackage{natbib}
\usepackage{graphicx}

\nocopyright

\begin{document}
%
\title{Gender Slopes: Counterfactual Fairness for Computer Vision Models \\ by Attribute Manipulation}


 \author{Jungseock Joo\thanks{jjoo@comm.ucla.edu}\\
{UCLA}\\
\And
Kimmo  K\"arkk\"ainen\\
{UCLA}
}


\maketitle

\begin{abstract}
Automated computer vision systems have been applied in many domains including security, law enforcement, and personal devices, but recent reports suggest that these systems may produce biased results, discriminating against people in certain demographic groups. Diagnosing and understanding the underlying true causes of model biases, however, are challenging tasks because modern computer vision systems rely on complex black-box models whose behaviors are hard to decode. We propose to use an encoder-decoder network developed for image attribute manipulation to synthesize facial images varying in the dimensions of gender and race while keeping other signals intact. We use these synthesized images to measure counterfactual fairness of commercial computer vision classifiers by examining the degree to which these classifiers are affected by gender and racial cues controlled in the images, e.g., feminine faces may elicit higher scores for the concept of nurse and lower scores for STEM-related concepts. We also report the skewed gender representations in an online search service on profession-related keywords, which may explain the origin of the biases encoded in the models. 
\end{abstract}


\section{Introduction}
Artificial Intelligence
has made remarkable progress in the past decade. Numerous AI-based products have already become prevalent in the market, ranging from robotic surgical assistants to self-driving vehicles. The accuracy of AI systems has surpassed human capability in challenging tasks, such as face recognition~\citep{taigman2014deepface}, lung cancer screening~\citep{ardila2019end} and pigmented skin lesion diagnosis~\citep{tschandl2019comparison}. These practical applications of AI systems have prompted  attention and support from industry, academia, and government. 

\begin{figure}
\centering
    \includegraphics[width=0.5\textwidth]{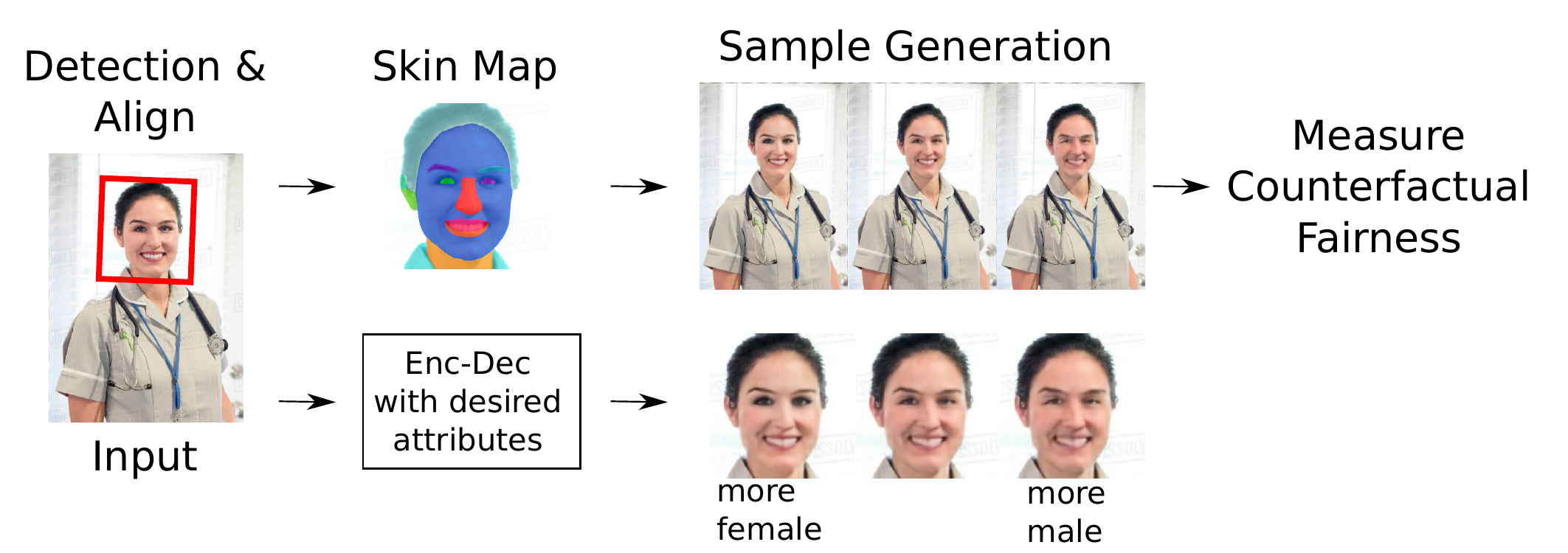}
\caption{Overview of our method for counterfactual image synthesis. }
\label{fig:algo}
\end{figure}

While AI technologies have contributed to increased work productivity and efficiency, a number of reports have also been made on the algorithmic biases and discrimination caused by data-driven decision making in AI systems. For example, COMPAS, an automated risk assessment tool used in criminal justice~\citep{brennan2009evaluating}, was reported to contain bias against Black defendants by assigning higher risk scores to Black defendants than White defendants~\citep{angwin_larson_kirchner_mattu_2019}. 
Another recent study also reports the racial and gender bias in computer vision APIs for facial image analysis, which were shown less accurate on certain race or gender groups~\citep{buolamwini2018gender}. 

How can biased machine learning and computer vision models impact our society? We consider a following example. Let's suppose an online search engine, such as Google, tries to make a list of websites of medical clinics and sort them by relevance. This list may be given to users as a search result or advertising content. The search algorithm will use content in websites to determine and rank their relevance, and any visual content, such as portraits of doctors, may be used as a feature in the pipeline. If the system relies on a biased computer vision model in this pipeline, the overall search results may also inherent the same biases and eventually affect users' decision makings. Scholars have discussed and found present biases in online media such as skewed search results~\citep{goldman2008search} or gender difference in STEM career ads
~\citep{lambrecht2019algorithmic}, yet little has been known about mechanisms or origins of such biases. 

While previous reports have shown that popular computer vision and machine learning models contain biases and exhibit disparate accuracies on different subpopulations, it is still difficult to identify true causes of these biases. This is because one cannot know to which variable or factor the model responds. If we wish to verify if a model indeed discriminates against a sensitive variable, e.g., gender, we need to isolate the factor of gender and intervene its value for \textbf{counterfactual} analysis~\citep{hardt2016equality}. 

The objective of our paper is to adopt an encoder-decoder architecture for facial attribute manipulation~\citep{lample2017fader} and generate counterfactual images which vary along the dimensions of sensitive attributes: gender and race. These synthesized examples are then used to measure counterfactual fairness of black-box image classifiers offered by commercial providers. Figure~\ref{fig:algo}  shows the overall process of our approach. Given an input image, we detect a face and generate a series of novel images by manipulating the target sensitive attributes while maintaining other attributes. 
We summarize our main contributions as follows.

\begin{figure*}
\centering
    \includegraphics[width=1.0\textwidth]{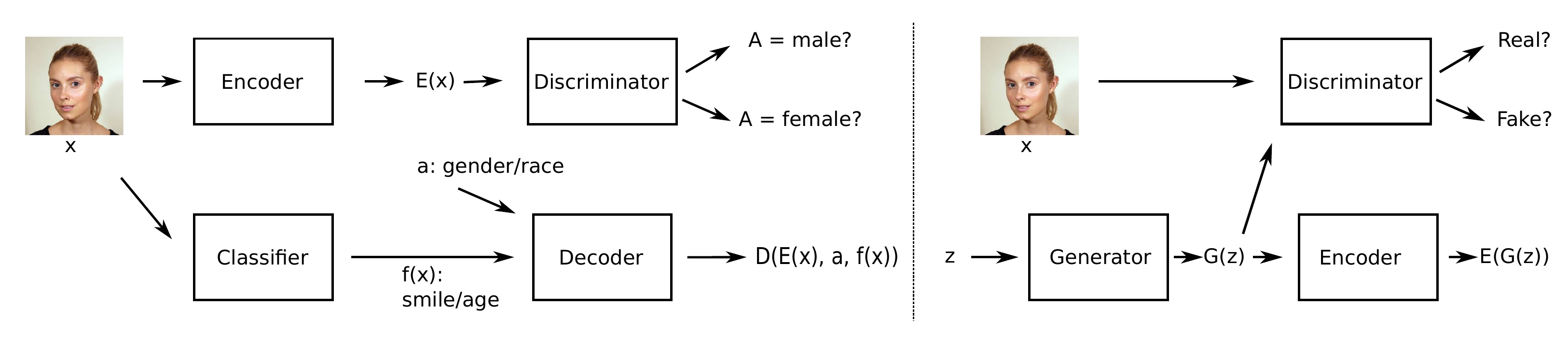}
    \caption{Illustrations of (left) our encoder-decoder architecture based on FaderNetwork~\citep{lample2017fader} and (right) a GAN used by Denton et al.~\citep{denton2019detecting}. Our model explicitly separates the sensitive attributes from the remaining representation encoded in $E(x)$.  In both models, the discriminator is optimized by adversarial training.}
\label{fig:faderarchi}
\end{figure*}

\begin{enumerate}
    \item We propose to use an encoder-decoder network~\citep{lample2017fader} to generate novel face images, which allows counterfactual interventions. Unlike previous methods~\citep{denton2019detecting}, our method explicitly isolates the factors for  sensitive attributes, which is critical in identifying true causes to model biases. 
    \item We construct a novel image dataset which consists of 64,500 original images collected from web search and more than 300,000 synthesized images manipulated from the original images. These images describe people in diverse occupations and can be used for studies on bias measurement or mitigation. Both the code and data will be made publicly available. 
    \item Using new methods and data, we measure counterfactual fairness of commercial computer vision classifiers and report whether and how sensitive these classifiers are affected along with attributes being manipulated by our model. 
\end{enumerate}




\section{Related Work}
\textbf{ML and AI Fairness}
Fairness in machine learning has recently received much attention as a new criterion for model evaluation~\citep{zemel2013learning,hardt2016equality,zafar2017fairness,kilbertus2017avoiding,kusner2017counterfactual}. While the quality of a machine learning model has traditionally been assessed by its overall performance such as average classification accuracy measured from the entire dataset, the new fairness measures focus on the consistency of model behavior across distinct data segments or the detection of spurious correlations between target variables (e.g., loan approval) and protected attributes (e.g., race or gender). 


The existing literature identifies a number of definitions and measures for ML/AI fairness~\citep{corbett2018measure}, including fairness through unawareness~\citep{dwork2012fairness}, disparate treatment and disparate impact~\citep{zafar2017fairness}, accuracy disparity~\citep{buolamwini2018gender}, and equality in opportunity~\citep{hardt2016equality}. These are necessary because different definitions of fairness should be used in different tasks and contexts. 

A common difficulty in measuring fairness is that it is challenging to identify or differentiate true causes of the discriminating model behaviors due to the input data that is built upon combination of many factors. Consequently, it is difficult to conclude that the variations in model outputs are solely caused by the sensitive or protected attributes. To overcome the limitation, Kusner et al.~\citep{kusner2017counterfactual} proposed the notion of counterfactual fairness based on causal inference. Here, a model, or predictor, is counterfactually fair as long as it produces an equal output to any input data whose values for the sensitive attribute are modified by an intervention but otherwise identical. 
Similar to \citep{kusner2017counterfactual}, our framework is based on counterfactual fairness to measure whether the prediction of the model differs by the \textit{intervened} gender of the input image, while separating out the influences from all the other factors in the background. 


\noindent
\textbf{Fairness and Bias in Computer Vision}
Fairness in computer vision is becoming more critical as many systems are being adapted in real world applications. For example, face recognition systems such as Amazon's Rekognition are being used by law enforcement to identify criminal suspects~\citep{harwell2019oregon}.  If the system produces biased results (e.g., higher false alarm on Black suspects), then it may lead to a disproportionate arrest rate on certain demographic groups.
In order to address this issue, scholars have attempted to identify biased representations of gender and race in public image dataset and computer vision models~\citep{hendricks2018women,manjunatha2019explicit,karkkainen2019fairface,mcduff2019characterizing}. Buolamwini and Gebru~\citep{buolamwini2018gender} have shown that commercial computer vision gender classification APIs are biased and thus perform least accurately on dark-skinned female photographs. \citep{kyriakou2019fairness} has also reported that image classification APIs may produce different results on faces in different gender and race. These studies, however, used the existing images without interventions, and thus it is difficult to identify whether the classifiers responded to the sensitive attributes or to the other visual cues. \citep{kyriakou2019fairness} used the headshots of people with clean white background, but this hinders the classifiers from producing many comparable tags. 

Our paper is most closely related to Denton et al.~\citep{denton2019detecting}, who use a generative adversarial network (GAN)~\citep{goodfellow2014generative} to generate face images to measure counterfactual fairness. Their framework incorporates a GAN trained from a face image dataset called CelebA~\citep{liu2015deep}, and generates a series of synthesized samples by modifying the latent code in the embedding space to the direction that would increase the strength in a given attribute (e.g., smile). 
Our paper differs from this work for the following reasons. First, we use a different method to examine the essential concept of counterfactual fairness by generating samples that separate the signals of the sensitive attributes out from the rest of the images. Second, our research incorporates the generated data to measure the bias of black-box image classification APIs whereas \citep{denton2019detecting} measures the bias of a dataset open to public~\citep{liu2015deep}. Using our distinct method and data, we aim to identify the internal biases of models trained from unknown data. 





\section{Counterfactual Data Synthesis}
\subsection{Problem Formulation}
The objective of our paper is to measure counterfactual fairness of a predictor $Y$, a function of an image $x$. This predictor is an image classifier that automatically labels the content of input images. Without the loss of generality, we consider a binary classifier, $Y(x) = \{True, False\}.$ This function classifies, for example, whether the image displays a doctor or not. 
We also define a sensitive attribute, $A$, gender and race. Typically, $A$ is a binary variable in the training data, but it can take a continuous value in our experiment since we can manipulate the value without restriction.  Following \citep{hardt2016equality}, this predictor satisfies counterfactual fairness if $P(Y_{A \leftarrow a} (x) = y| x ) = P(Y_{A \leftarrow a'} (x) = y| x )$ for all $y$ and any $a$ and $a'$,
where $A \leftarrow a$ indicates an intervention on the sensitive attribute, $A$. We now explain how this is achieved by an encoder-detector network. 

The goal of this intervention is to manipulate an input image such that it changes the cue related to the sensitive attribute while retaining all the other signals. We consider two sensitive attributes: gender and race. We manipulate facial appearance because face is the strongest cue for gender and race identification~\citep{moghaddam2002learning}.

\begin{figure}
\centering
    \includegraphics[width=0.5\textwidth]{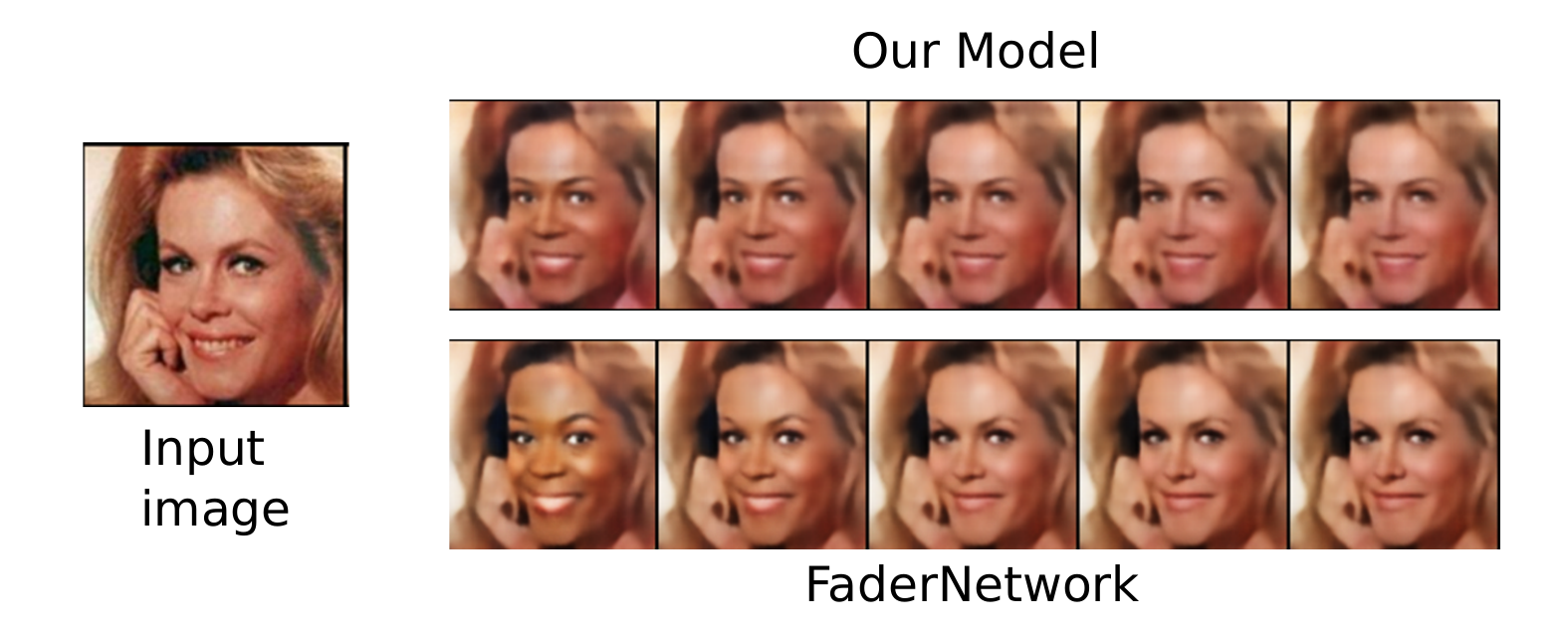}
    \caption{Our model controls for non-central attributes such as smiling and age. These attributes (e.g., mouth open) are fixed while the main attribute (race) is manipulated.  }
\label{fig:controlattribute}
\end{figure}

\subsection{Counterfactual Data Synthesis}
Before we elaborate our proposed method for manipulating sensitive attributes, we briefly explain why such a method is necessary to show if a model achieves counterfactual fairness. For an in-depth introduction to the framework of counterfactual fairness, we refer the reader to Kusner et al.~\citep{kusner2017counterfactual}.  

Many studies have reported skewed classification accuracy of existing computer vision models and APIs between gender and racial groups~\citep{buolamwini2018gender,kyriakou2019fairness,karkkainen2019fairface,zhao2017men}. However, these findings are based on a comparative analysis, which directly compares the classifier outputs between male and female images (or White and non-White) in a given dataset. The limitation of the method is that it is difficult to identify true sources of biased model outputs due to hidden confounding factors. Even though one can empirically show differences between gender groups, such differences may have been caused by non-gender cues such as hair style or image backgrounds (see \citep{muthukumar2018understanding}, for example). Since there exists an infinite number of possible confounding factors, it will be very difficult to control for all of them. 

Consequently, recent works in bias measurement or mitigation have adopted generative models which can synthesize or manipulate text or image data \citep{denton2019detecting, zmigrod2019counterfactual}. These methods generate hypothetical data in which only sensitive attributes are switched. These data can be used to measure counterfactual fairness but also augment samples in existing biased datasets.

\begin{figure*}
\centering
    \includegraphics[width=0.95\textwidth]{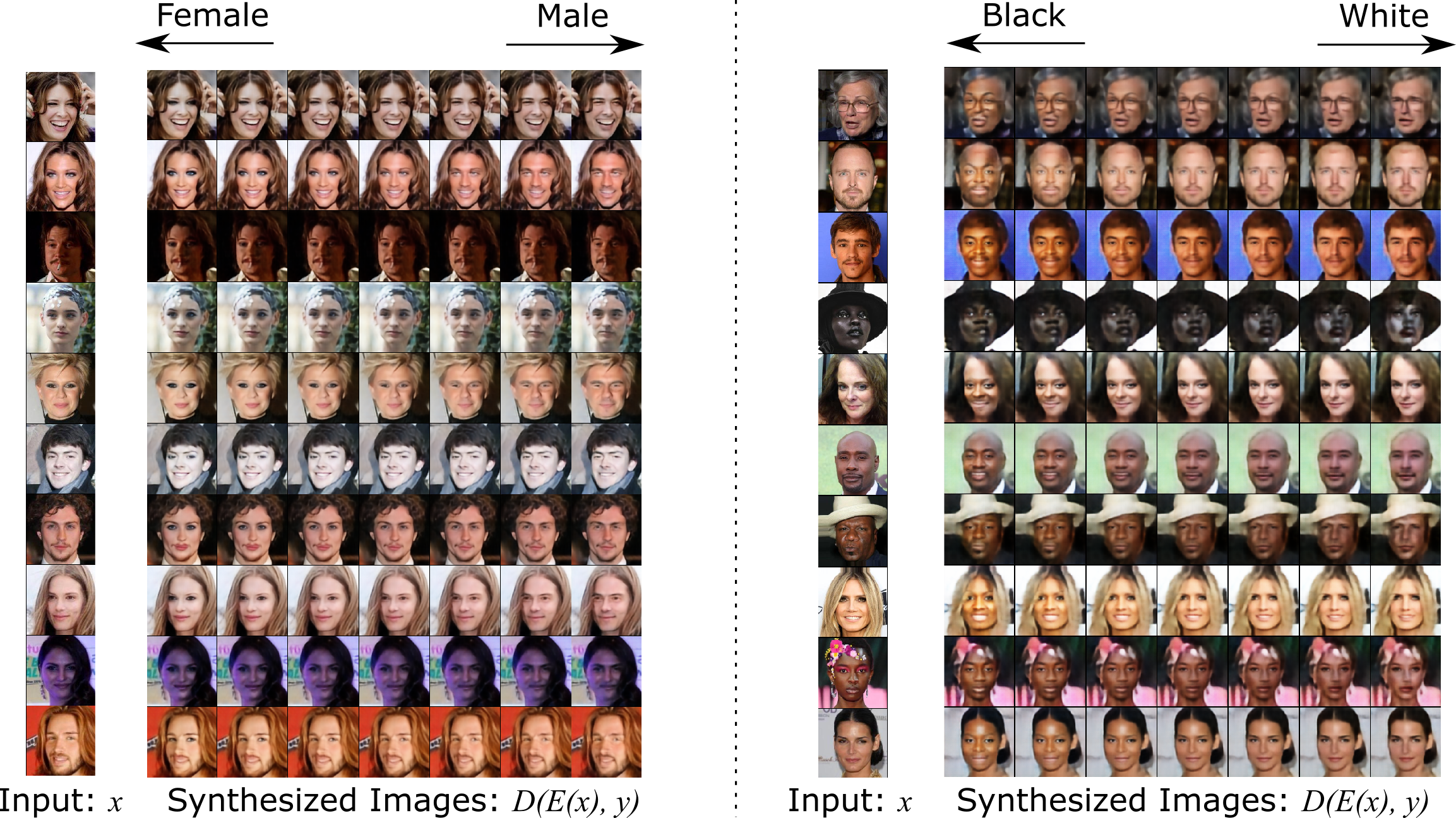}
    \caption{Examples generated by our models, manipulated in (left) gender and (right) race. }
\label{fig:faderexample}
\end{figure*}

\subsection{Face Attribute Synthesis}
From the existing methods available for face attribute manipulation~\citep{yan2016attribute2image, bao2017cvae, he2019attgan}, we chose FaderNetwork~\citep{lample2017fader} as our base model. FaderNetwork is a computationally efficient model that produces plausible results, but we made a few changes to make it more suitable for our study. 

Figure~\ref{fig:faderarchi} illustrates the flows of our model and \citep{denton2019detecting}. The model used in \citep{denton2019detecting} is based on a GAN that is trained without using any attribute labels. As in standard GANs, this model learns the latent code space from the training set. This space encodes various information such as gender, age, race, and any other cues necessary for generating a facial image. These factors are all entangled in the space, and thus it is hard to control only the sensitive attribute, which is required for the purpose of counterfactual fairness measurement. In contrast, FaderNetwork directly observes and exploits the sensitive attributes in training and makes its latent space invariant to them. 

Specifically, FaderNetwork is based on an encoder-decoder network with two special properties. First, it separates the sensitive attribute, $a$, from its encoder output, $E(x)$, and both are fed into the decoder, such that it can reconstruct the original image, i.e., $D(E(x), a) \approx x$. Second, it makes $E(x)$ invariant to $a$ by using adversarial training such that the discriminator cannot predict the correct value for $a$ given $E(x)$. At test time, an arbitrary value for $a$ can be given to obtain an image with a modified attribute value. 

Since we want to minimize the change by the model to dimensions other than the sensitive attributes, we added two additional steps as follows. First, we segment the facial skin region from an input face by~\citep{yu2018bisenet}\footnote{https://github.com/zllrunning/face-parsing.PyTorch} and only retain changes within the region. This prevents the model from affecting background or hair regions. Second, we control for the effects of other attributes (e.g., smiling or young) which may be correlated with the main sensitive attribute, such that their values remain intact while being manipulated. This was achieved by first modeling these attributes as the main sensitive attributes along with $y$ in training and fixing their values at testing time. 
This step may look unnecessary because the model is expected to separate all gender (or any other sensitive attributes) related information. However, it is important to note that the dataset used to train our model may also contain biases and it is hard to guarantee that its sensitive attributes are not correlated with other attributes. By enforcing the model to produce fixed outputs, we can explicitly control for those variables (similar ideas have been used in recent work on attribute manipulation \citep{he2019attgan}). 
Figure~\ref{fig:controlattribute} shows the comparison between our model and the original FaderNetwork. 
This approach allows our model to minimize the changes in dimensions other than the main attribute being manipulated.  
Figure~\ref{fig:faderexample} shows randomly chosen results by our method.


\section{Experiments}
\subsection{Computer Vision APIs} 
We measured counterfactual fairness of commercial computer vision APIs which provide label classification for a large number of visual concepts, including Google Vision API, Amazon Rekognition, IBM Watson Visual Recognition, and Clarifai. These APIs are widely used in commercial products as well as academic research~\citep{xi2019understanding}. While public computer vision datasets usually focus on general concepts (e.g., 60 common object categories in MS COCO~\citep{lin2014microsoft}), these services generate very specific and detailed labels on thousands of distinct concepts. While undoubtedly useful, these APIs have not been fully verified for their fairness.  They may be more likely to generate more ``positive'' labels for people in certain demographic groups. These labels may include highly-paid and competitive occupations such as ``doctor'' or ``engineer'' or personal traits such as ``leadership'' or ``attractive''.  We measure the sensitivity of these APIs using counterfactual samples generated by our models.

\subsection{Occupational Images}
We constructed the baseline data that can be used to synthesize samples. We are especially interested in the effects of gender and race changes on the profession related labels provided by the APIs, and thus collected a new dataset of images related to various professions. 
We first obtained a list of 129 job titles from the Bureau of Labor Statistics (BLS) website and used Google Image search to download images. Many keywords resulted in biased search results in terms of the gender and race ratio. To obtain more diverse images, we additionally combined six different keywords (male, female, African American, Asian, Caucasian, and Hispanic). This results in around 250 images per keyword. We disregarded images without any face.

\begin{figure*}[t]
\centering
     \begin{subfigure}[b]{0.45\textwidth}
        \centering
        \includegraphics[width=\textwidth]{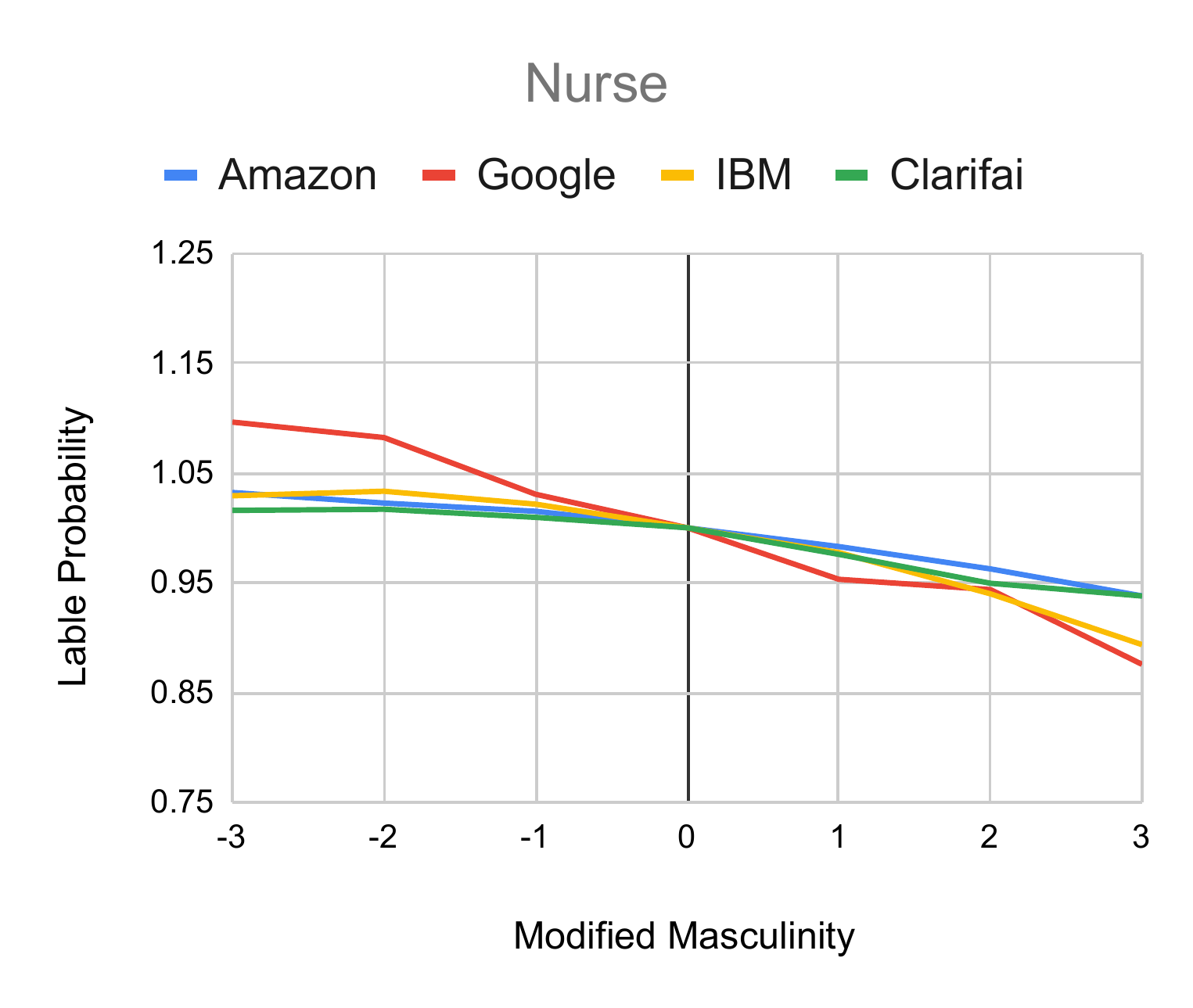}
     \end{subfigure}   ~~~~~~~~~~
     \begin{subfigure}[b]{0.45\textwidth}
        \centering
        \includegraphics[width=\textwidth]{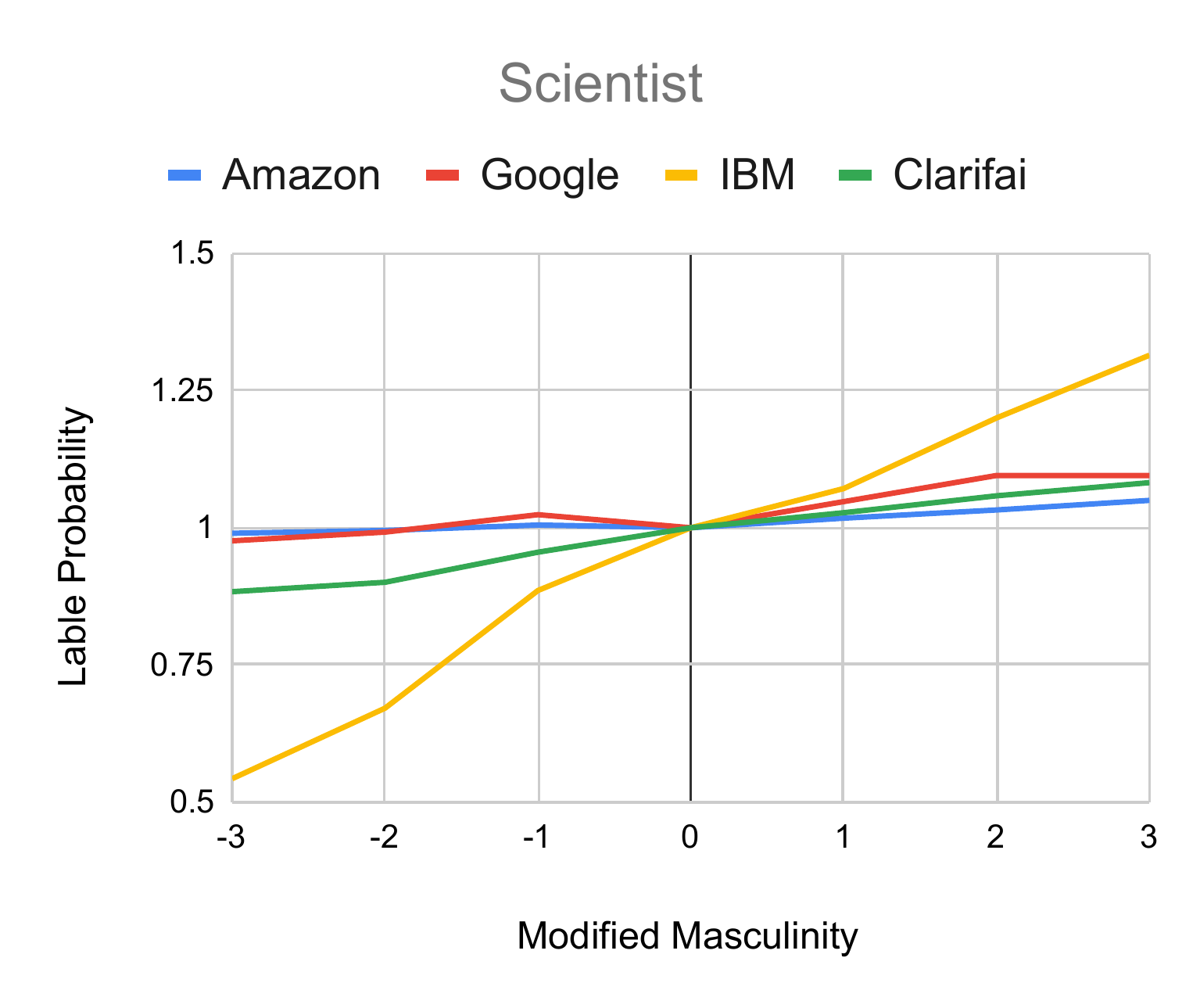}
     \end{subfigure}
\caption{The sensitivity of image classification APIs for Nurse and Scientist to the modified facial gender cues. }
\label{fig:nurse}
\end{figure*}

We also needed datasets for training our model. For the gender manipulation model, we used CelebA~\citep{liu2015deep}, which is a very popular face attribute dataset with 40 labels annotated for each face.  This dataset mostly contains the faces of White people, and thus is not suitable for the race manipulation model. There is no publicly available dataset with a sufficiently large number of African Americans. Instead, we obtained the list of the names of celebrities for each gender and each ethnicity from an online website, FamousFix. Then we used Google Image search to download up to 30 images for each celebrity. We estimated the true gender and race of each face by a model trained from a public dataset~\citep{karkkainen2019fairface} and manually verified examples with lower confidences. Finally, this dataset was combined with CelebA to train the race manipulation model. 

After training, two models (gender and race) were applied to the profession dataset to generate a series of manipulated images for each input image. If there are multiple faces detected in an image, we only manipulated the face closest to the center of it. 
These faces are pasted into the original image, only on the facial skin region, and passed to each of the 4 APIs we tested. All the APIs provide both the presence of each label (binary) and the continuous classification confidence if the concept is present in the image.  
Figure~\ref{fig:faderexample} shows example images manipulated in gender and race. 


\begin{figure*}
\centering
    \includegraphics[width=0.95\textwidth]{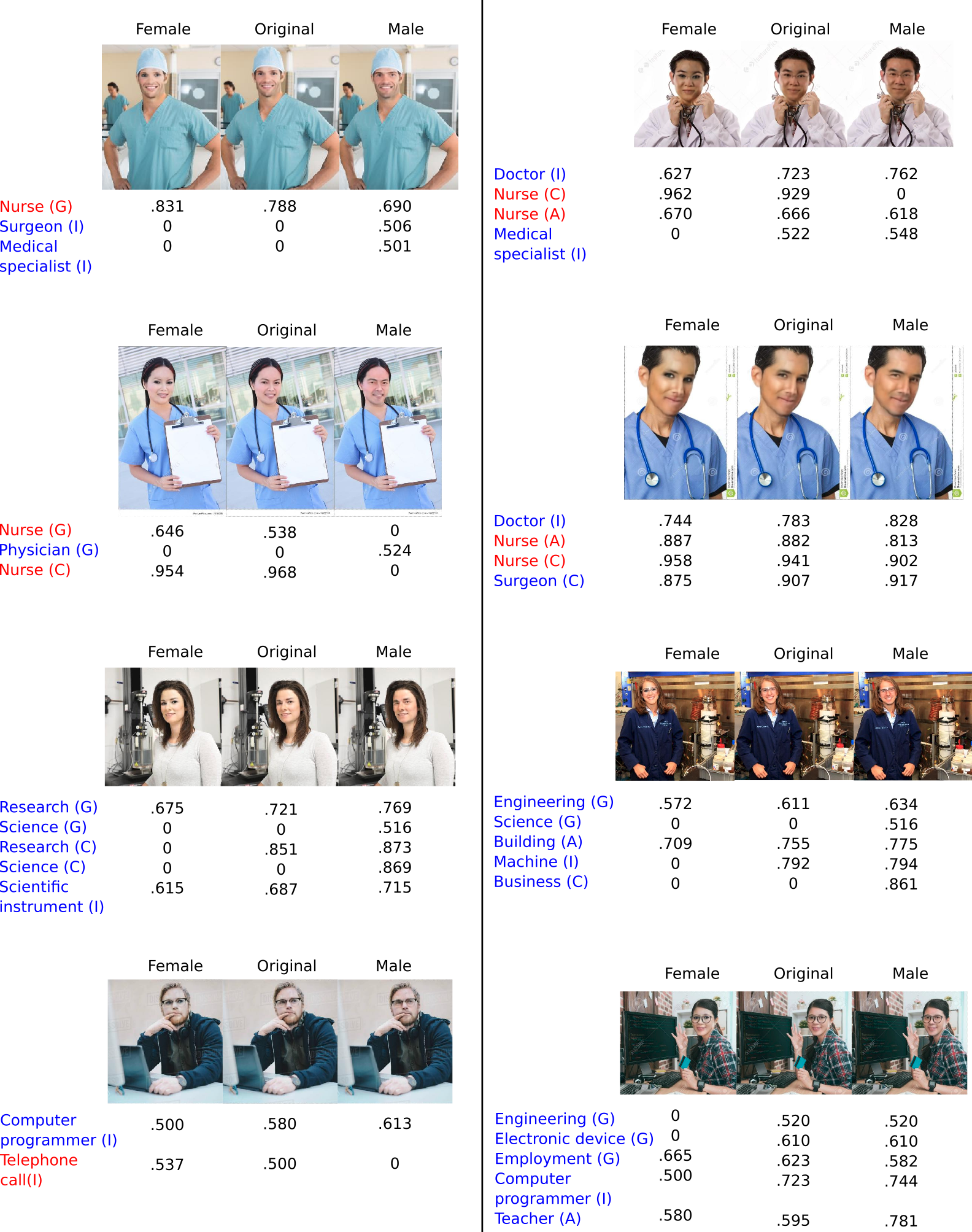}
    \caption{Example images and label prediction scores from APIs (G:Google, A:Amazon, I:IBM, C:Clarifai). ``0'' means the label was not detected. Blue labels indicate an increasing score with increasing masculinity (red for femininity). Some images were clipped to fit the space.  Zoom in to see the details. }
\label{fig:qualexample}
\end{figure*}

\subsection{Results}
The sensitivity of a classifier with respect to the changes in gender or race cues of images is measured as a slope estimated from the assigned attribute value, $a$, and the model output, $Y( x(a))$, where $x(a)$ is a synthesized image with its attribute manipulated to the value $a$. The range of $a$ was set to $(-2, 2)$. The center, i.e., gender-neutral face, is 0. $(-1, 1)$ is the range observed in training, and $(-2, 2)$ will extrapolate images beyond the training set. In practice, this still results in natural and plausible samples. From this range, we sampled 7 evenly spaced images for gender manipulation and 5 images for race manipulation.\footnote{We reduced the number from 7 to 5 as this was more cost effective and sufficient to discover the correlation between the attributes and output labels.  } Let us denote $x^i$, the $i$-th input image, and $\{x^i_1, .., x^i_K\}$, the set of $K$ synthesized images ($K = 7$). For each label in $Y$, we obtain 7 scores. From the entire image set $\{x^i\}$, we obtain a normalized classifier output vector:
\begin{equation*}
\begin{aligned}
    y_k &= \frac{1}{n} \sum_i{\mathbb{1} \{ Y(x^i_k) = True \}}, k \in \{1, .., K\}, \\
    z_k &=  y_k / y_c, c = (K+1) / 2.
\end{aligned}
\end{equation*}
That is, we normalize the vector such that $z_c$ is always 1 to allow comparisons across concepts. The slope $b$ is obtained by linear regression with ordinary least squares. The magnitude of $b$ determines the sensitivity of the classifier against $a$, and its sign indicates the direction. 

Table \ref{table:slopefemale} and \ref{table:slopemale} show the list of labels returned by each API, more frequently activated with images manipulated to be closer to women and to men, respectively. Not surprisingly, we found the models behave in a closely related way to the actual gender gap in many occupations such as nurses or scientists (see Figure \ref{fig:nurse}, too). One can imagine this bias was induced at least in part due to the bias in the online media and web, from which the commercial models have been trained. 
Table~\ref{table:gendersearch} and \ref{table:racegoogle} show skewed gender and race representations in our main dataset of peoples' occupations. Indeed, many occupations such as nurse or engineer exhibit very sharp gender contrast, and this may explain the behaviors of the image classifiers.  Figure~\ref{fig:qualexample} shows example images and their label prediction scores.\footnote{The APIs output a binary decision and a prediction confidence for each label. Our analysis is based on binary values (true or false), and we found that using confidence scores makes little difference in the final results. } 

Similarly, Table~\ref{tab:raceconcept} and \ref{tab:raceconceptwhite}
 show the labels which are most sensitive to the race manipulation. The tables show all the dimensions which are significantly correlated with the model output ($p<0.001$), except plain concepts such as ''Face'' or ''Red color''. We found the APIs are in general less sensitive to race change than gender change.

\begin{table}[]
\caption{The Sensitivity of Label Classification APIs against Gender Manipulation (Female). (Only showing labels with p-value $<$ 0.001 and $|$ slope $|$ $>$ 0.03). }
\begin{tabular}{llc}
API	&	Label	&	Slope	\\ \hline
Amazon	&	Nurse	&	-0.031	\\ \hline
Google	&	Fashion model	&	-0.262	\\
Google	&	Model	&	-0.261	\\
Google	&	Secretary	&	-0.14	\\
Google	&	Nurse	&	-0.073	\\ \hline
IBM	&	anchorperson	&	-0.213	\\
IBM	&	television reporter	&	-0.155	\\
IBM	&	college student	&	-0.151	\\
IBM	&	legal representative	&	-0.147	\\
IBM	&	careerist	&	-0.128	\\
IBM	&	host	&	-0.125	\\
IBM	&	steward	&	-0.11	\\
IBM	&	Secretary of State	&	-0.107	\\
IBM	&	gynecologist	&	-0.099	\\
IBM	&	celebrity	&	-0.097	\\
IBM	&	newsreader	&	-0.09	\\
IBM	&	cleaning person	&	-0.081	\\
IBM	&	nurse	&	-0.046	\\
IBM	&	laborer	&	-0.044	\\
IBM	&	workman	&	-0.041	\\
IBM	&	entertainer	&	-0.04	\\ \hline
Clarifai	&	secretary	&	-0.273	\\
Clarifai	&	receptionist	&	-0.268	\\
Clarifai	&	model	&	-0.211	  \\
Clarifai	&	shopping	&	-0.058	  \\
\end{tabular}%
\label{table:slopefemale}
\end{table}


\begin{table}[]
\caption{The Sensitivity of Label Classification APIs against Gender Manipulation (Male). (Only showing labels with p-value $<$ 0.001 and $|$ slope $|$ $>$ 0.03).}
\begin{tabular}{llc}
API    & Label                     & Slope \\ \hline
Amazon	&	Attorney		& .113	\\ 
Amazon	&	Executive		& .055	\\ \hline
Google	&	Blue-collar worker		& .056	\\ 
Google	&	Spokesperson		& .040	\\
Google	&	Engineer		& .038	\\ \hline
IBM	& scientist	& .254			\\ 
IBM	& sociologist	& .213			\\
IBM	& investigator	& .174			\\
IBM	& sports announcer	& .164			\\
IBM	& resident commissioner	& .159			\\
IBM	& repairer	& .151			\\
IBM	&	Representative		& .142	\\
IBM	&	cardiologist		& .140	\\
IBM	&	high commissioner		& .134	\\
IBM	&	security consultant		& .131	\\
IBM	&	speaker		& .122	\\
IBM	&	internist		& .119	\\
IBM	&	Secretary of the Int.		& .114 	\\
IBM	&	biographer		& .109	\\
IBM	&	military officer		& .107	\\
IBM	&	radiologist		& .082	\\
IBM	&	detective		& .081	\\
IBM	&	diplomat		& .063	\\
IBM	&	contractor		& .061	\\
IBM	&	player		& .061	\\
IBM	&	medical specialist		& .050	\\
IBM	&	official		& .049	\\
IBM	&	subcontractor		& .043	\\ \hline
Clarifai	& film director	& .342	\\
Clarifai	& machinist	& .192			\\
Clarifai	& writer	& .153			\\
Clarifai	& repairman	& .125			\\
Clarifai	& surgeon	& .087			\\
Clarifai	& inspector	& .085  			\\
Clarifai	& waiter	& .082			\\
Clarifai	& worker	& .078			\\
Clarifai	& scientist	& .070			\\
Clarifai	& singer	& .056			\\
Clarifai	& musician	& .056			\\
Clarifai	& construction worker	& .053			\\
Clarifai	& police	& .054			\\
Clarifai	& athlete	& .048			\\
Clarifai	& politician	& .037			\\
\end{tabular}%
\label{table:slopemale}
\end{table}


\begin{table}[]
\caption{Skewed gender representations in Google Image search result}
\label{table:gendersearch}
\resizebox{0.5\textwidth}{!}{%
\begin{tabular}{lc|lc}
Occupation                     & Female \% & Occupation              & Male \% \\ \hline
nutritionist                   & .921      & pest control worker     & .971    \\
flight attendant               & .891      & handyman                & .964    \\
hair stylist                   & .884      & logging worker          & .950    \\
nurse                          & .860      & basketball player       & .925    \\
medical assistant              & .847      & businessperson          & .920    \\
dental assistant               & .835      & chief executive officer & .917    \\
merchandise displayer          & .821      & lawn service worker     & .909    \\
nursing assistant              & .821      & electrician             & .901    \\
dental hygienist               & .815      & barber                  & .901    \\
veterinarian                   & .784      & repair worker           & .900    \\
fashion designer               & .775      & sales engineer          & .889    \\
occupational therapy asst.  & .772      & construction worker     & .887    \\
libarian                       & .770      & maintenance worker      & .882    \\
office assistant               & .759      & officer                 & .882    \\
receptionist                   & .745      & radio operator          & .871    \\
travel agent                   & .734      & music director          & .868    \\
medical transcriptionist       & .732      & software developer      & .857    \\
preschool teacher              & .730      & golf player             & .855    \\
teacher assistant              & .728      & CTO                     & .846    \\
counselor                      & .728      & mechanic                & .836   
\end{tabular}%
}
\end{table}

\begin{table}[]
\caption{Skewed race representations in Google Image search result}
\label{table:racegoogle}
\resizebox{0.5\textwidth}{!}{%
\begin{tabular}{lc|lc}
Occupation                    & White \% & Occupation                        & White \% \\ \hline
historian              & .885     & basketball player          & .200     \\
building inspector     & .875     & farmworker                 & .231     \\
funeral director       & .852     & ahtlete                    & .415     \\
construction inspector & .847     & software developer         & .429     \\
glazier                & .846     & product promoter           & .451     \\
legislator             & .840     & interpreter                & .457     \\
animal trainer         & .839     & barber                     & .457     \\
boiler installer       & .836     & medical assistant          & .459     \\
jailer                 & .823     & food scientist             & .462     \\
judge                  & .822     & chemical engineer          & .488     \\
handyman               & .821     & database administrator     & .489     \\
baker                  & .818     & computer network architect & .500     \\
firefighter            & .815     & industrial engineer        & .513     \\
veterinarian           & .811     & driver                     & .514     \\
pilot                  & .797     & bus driver                 & .532     \\
optician               & .795     & fashion designer           & .539     \\
businessperson         & .793     & security guard             & .539     \\
CFO                    & .791     & mechanic                   & .548     \\
maintenance manager    & .791     & nurse                      & .548     \\
secretary              & .785     & cashier                    & .549    
\end{tabular}%
}
\end{table}

\begin{table}[]
\caption{The Sensitivity of Label Classification APIs against Race Manipulation (Black). (Only showing labels with p-value $<$ 0.001 and $|$ slope $|$ $>$ 0.03).}
\label{tab:raceconcept}
\begin{tabular}{ccc}
API      & Label                          & Slope \\ \hline 
IBM	&	woman orator			&	-0.69	\\
IBM	&	President of the U.S.		&	-0.367	\\
IBM	&	first lady			&	-0.323	\\
IBM	&	high commissioner			&	-0.284	\\
IBM	&	Representative			&	-0.225	\\
IBM	&	scientist			&	-0.183	\\
IBM	&	worker			&	-0.131	\\
IBM	&	resident commissioner			&	-0.116	\\
IBM	&	sociologist			&	-0.099	\\
IBM	&	analyst			&	-0.09	\\
IBM	&	call center			&	-0.085	\\
IBM	&	diplomat			&	-0.085	\\ \hline
Clarifai	&	democracy			&	-0.09	\\
Clarifai	&	musician			&	-0.063	\\
Clarifai	&	singer			&	-0.046	\\
Clarifai	&	cheerful			&	-0.044	\\
Clarifai	&	happiness			&	-0.034	\\
Clarifai	&	music			&	-0.033	\\
Clarifai	&	confidence			&	-0.032	\\ 
\end{tabular}%
\end{table}

\begin{table}[]
\caption{The Sensitivity of Label Classification APIs against Race Manipulation (White). (Only showing labels with p-value $<$ 0.001 and $|$ slope $|$ $>$ 0.03).}
\label{tab:raceconceptwhite}
\begin{tabular}{ccc}
API      & Label                          & Slope \\ \hline
IBM	&	careerist	&	0.179		\\
IBM	&	dermatologist (doctor)	&	0.127		\\
IBM	&	legal representative	&	0.111		\\
IBM	&	business man	&	0.093		\\
IBM	&	entertainer	&	0.034		\\ \hline
Clarifai	&	repair	&	0.074		\\
Clarifai	&	beautiful	&	0.074		\\
Clarifai	&	repairman	&	0.073		\\
Clarifai	&	writer	&	0.054		\\
Clarifai	&	physician	&	0.053		\\
Clarifai	&	work	&	0.051		\\
Clarifai	&	professional person	&	0.05		\\
Clarifai	&	contractor	&	0.05		\\
Clarifai	&	fine-looking	&	0.048		\\
Clarifai	&	skillful	&	0.044		\\
Clarifai	&	pretty	&	0.039		\\ \hline
Google	&	Beauty	&	0.06		
\end{tabular}%
\end{table}

\section{Conclusion}
AI fairness is an increasingly important criterion to evaluate models and systems. In real world applications, especially for private models whose training processes or data are unknown, it is difficult to identify their biased behaviors or to understand the underlying causes. We introduced a novel method based on facial attribute manipulation by an encoder-decoder network to synthesize counterfactual samples, which can help isolate the effects of the main sensitive variables on the model outcomes. Using this methodology, we were able to identify hidden biases of commercial computer vision APIs on gender and race. These biases, likely caused by the skewed representation in online media, should be adequately addressed in order to make these services more reliable and trustworthy. 

\section{Acknowledgement}
This work was supported by the National Science Foundation SMA-1831848.


\bibliographystyle{aaai}
\bibliography{sample-base}

\end{document}